\documentclass[conference]{IEEEtran}
\IEEEoverridecommandlockouts
\usepackage{cite}
\usepackage{amsmath,amssymb,amsfonts}
\usepackage{algorithmic}
\usepackage{graphicx}
\usepackage{subcaption}
\usepackage{textcomp}
\usepackage{float}
\usepackage{url}
\usepackage{xcolor}
\def\BibTeX{{\rm B\kern-.05em{\sc i\kern-.025em b}\kern-.08em
    T\kern-.1667em\lower.7ex\hbox{E}\kern-.125emX}}

\begin{document}

\title{\\
{\footnotesize \
}
}

\makeatletter
\newcommand{\newlineauthors}{%
  \end{@IEEEauthorhalign}\hfill\mbox{}\par
  \mbox{}\hfill\begin{@IEEEauthorhalign}
}
\makeatother

\title{InfoTech Assistant: A Multimodal Conversational Agent for InfoTechnology Web Portal Queries}

\author{
    \IEEEauthorblockN{Sai Surya Gadiraju, Duoduo Liao, Akhila Kudupudi, Santosh Kasula, Charitha Chalasani} \\
    \IEEEauthorblockA{
        \textit{School of Computing} \\
        \textit{George Mason University} \\
        \textit{Fairfax, VA USA} \\
        \textit{\{sgadira3, dliao2, akudpudi, skasula2,  cchalasa\}@gmu.edu} \\
    }
}

\maketitle

\begin{abstract}
This pilot study presents the development of the InfoTech Assistant, a domain-specific, multimodal chatbot engineered to address queries in bridge evaluation and infrastructure technology. By integrating web data scraping, large language models (LLMs), and Retrieval-Augmented Generation (RAG), the InfoTech Assistant provides accurate and contextually relevant responses. Data, including textual descriptions and images of 41 bridge technologies, are sourced from publicly available documents on the InfoTechnology website and organized in JSON format to facilitate efficient querying. The architecture of the system includes an HTML-based interface and a Flask back end connected to the Llama 3.1 model via LLM Studio. Evaluation results show approximately 95 percent accuracy on domain-specific tasks, with high similarity scores confirming the quality of response matching. This RAG-enhanced setup enables the InfoTech Assistant to handle complex, multimodal queries, offering both textual and visual information in its responses. The InfoTech Assistant demonstrates strong potential as a dependable tool for infrastructure professionals, delivering high accuracy and relevance in its domain-specific outputs.
\end{abstract}


\section{Introduction}

The rapid growth of information in infrastructure technology has created an increasing demand for advanced tools that streamline knowledge retrieval and dissemination. With the continuous expansion of data in fields such as engineering, construction, and maintenance, it has become increasingly difficult for professionals to efficiently access and utilize the vast amounts of knowledge available. The Federal Highway Administration (FHWA) InfoTechnology \cite{b0} platform which acts as a comprehensive repository, that consolidates a wide range of infrastructure-related information, spanning domains such as bridges, pavements, tunnels, and utilities. This platform is designed to support decision-making processes and enhance the effectiveness of infrastructure management by providing a centralized source of technical data, standards, guidelines, and research findings\cite{b101, b102}.

Our study specially focuses on bridge-related technologies, with an emphasis on information pertaining to inspection, assessment, and maintenance. While  the abundance of structured information within the platform is valuable, the sheer volume and complexity of content often presents challenges for users to locate specific information within its extensive content. 

To address these challenges, the pilot \textit{InfoTech Assistant} has been developed as an interactive multimodal conversational tool, leveraging Large Language Models (LLMs) and Natural Language Processing (NLP) techniques to to enhance information retrieval and user interaction. This system allows users to efficiently access relevant information through a conversational interface. By employing a Retrieval-Augmented Generation (RAG) approach \cite{b5}, the InfoTech Assistant integrates real-time data retrieval with language generation, ensuring precise, contextually relevant responses tailored to bridge technology.

The InfoTech Assistant comprises several key components, including data collection, a user-friendly interface \cite{b19}, and integration with a state-of-the-art LLM. Data collection employs automated web scraping techniques to extract textual and visual data from the InfoTechnology web portal \cite{b1}. The user interface allows users to submit queries and receive detailed answers, supplemented with relevant images. Integrated semantic retrieval capabilities ensure the assistant delivers precise, domain-specific responses, making it highly effective for infrastructure professionals and researchers.

The primary contributions are outlined as follows:

\begin{itemize}
     \item \textit{Conversational Agent with Multimodal Integration:} The InfoTech Assistant enhances response accuracy by integrating both text and image data from the InfoTechnology platform, providing contextual understanding and visual support that caters to the needs of technical professionals in infrastructure domains\cite{b24}.
    \item \textit{Domain-Specific Knowledge Retrieval Using RAG:} This system combines data retrieval with language generation capabilities, allowing for precise, adaptable responses optimized for complex, domain-specific queries in InfoTechnology.\cite{b5}
    \item \textit{User-Centric System Design for Real-Time Interaction:} The InfoTech Assistant’s architecture, featuring a structured JSON database, a responsive HTML interface, and a Flask-based back end with LLM integration, ensures seamless, real-time access to information. \cite{b6,b8}
    \item \textit{Quantitative Evaluation Framework for Response Quality:} Using cosine similarity and accuracy metrics, this evaluation framework rigorously assesses response relevance and precision, enhancing the reliability of the InfoTech Assistant in infrastructure-related queries.
    \item \textit{Domain-Specific Focus Tailored for Bridge Technology Professionals:} Designed specifically for professionals in bridge evaluation and infrastructure, the assistant delivers technical insights and comprehensive responses, supporting informed decision-making in the field. 
\end{itemize}

This pilot study details the development of the InfoTech Assistant, describing each component from data pre-processing to LLM integration and deployment. It also elaborates on the RAG framework \cite{b21}, which combines reliable data retrieval with language generation to produce accurate outputs. Finally, the paper discusses the application of the InfoTech Assistant in real-world infrastructure settings, emphasizing its role in providing quick and dependable access to critical information.

\begin{figure*}[h]
\centerline{\includegraphics[width=0.8\linewidth]{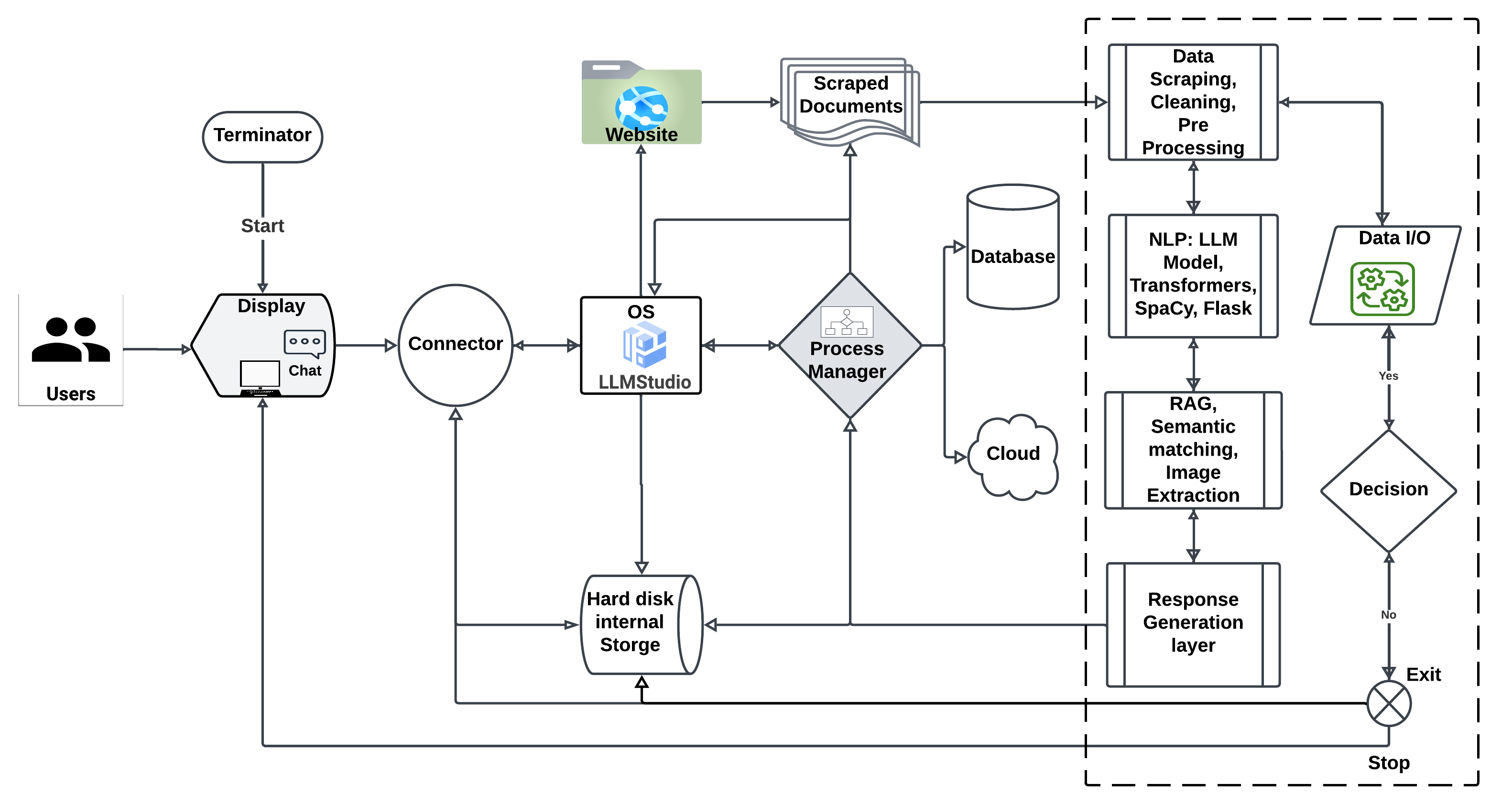}}
\caption{The System Architecture of the InfoTech Assistant}
\label{fig: Fig1}
\end{figure*}

\section{Related Work}
The InfoTech Assistant draws inspiration from both general-purpose and domain-specific Question Answering (QA) systems\cite{b15}, leveraging advancements in RAG\cite{b21}, multi-modal integration, and NLP techniques. By examining the strengths and limitations of these systems, the InfoTech Assistant integrates proven approaches while addressing challenges specific to infrastructure evaluation and Non-Destructive Evaluation (NDE) technologies. The following review outlines key contributions that informed the design and functionality of the InfoTech Assistant.
\vspace{-0.2cm}
\subsection{General-Purpose QA Systems}
General-purpose QA systems, such as IBM Watson, have demonstrated robust capabilities in natural language understanding and retrieval \cite{b9}. However, these systems often lack the domain-specific depth required for technical fields like NDE. Frameworks such as SearchQA \cite{b9} and DrQA \cite{b10} introduced techniques for semantic similarity and TF-IDF-based query alignment, which, while effective for open-domain QA, exhibited limited accuracy in handling specialized content. The proposed InfoTech Assistant builds on these approaches by utilizing semantic embeddings for precise query matching, thereby enhancing retrieval accuracy in domain-specific contexts.
\vspace{-0.2cm}
\subsection{Domain-Specific QA Systems}
BioASQ \cite{b11} highlights the successful application of Named Entity Recognition (NER) and domain-specific NLP for handling specialized corpora. This demonstrates the need for tailored approaches in technical domains. Similarly, Haystack \cite{b12} integrates RAG frameworks to deliver highly accurate and contextually grounded responses. These methodologies influenced the design of our InfoTech assistant by emphasizing the importance of combining semantic retrieval with generative modeling to address complex infrastructure-related queries effectively.
\vspace{-0.2cm}
\subsection{Chatbot Applications in Construction}
Applications like Skanska’s Sidekick \cite{b13} showcase the utility of chatbots in construction, providing natural language interfaces to access construction data. However, these tools focus on logistical queries and lack depth for technical evaluations. Our InfoTech Assistant extends these capabilities by incorporating detailed material-specific insights, making it suitable for NDE scenarios\cite{b35} where precise and contextual responses are critical.
\vspace{-0.2cm}
\subsection{Multi-Modal and RAG Integration}
Systems such as MS MARCO \cite{b14} and Visual QA \cite{b15} demonstrate the value of multi-modal data in enriching user interactions by integrating structured data with image-text retrieval. The InfoTech Assistant adopts similar methodologies by structuring its dataset in JSON format, enabling efficient retrieval of both textual and visual information. Inspired by large language models like Turing-NLG \cite{b16}, the Assistant leverages Llama 3.1 \cite{b7} to ensure scalability, secure query handling, and advanced generative capabilities.

\section{Methodology}

The development of the InfoTech Assistant follows a multi-phase methodology, encompassing data collection via web scraping \cite{b20}, front-end interface development, and back-end integration with a pre-trained large language model (LLM) to enable efficient information retrieval and user interaction.

\subsection{The System Architecture}

The architecture comprises several essential components, each fulfilling a vital role in enabling seamless data processing, precise information retrieval, and responsive user interaction, as illustrated in Fig. \ref{fig: Fig1}, which was created using Lucidchart \cite{b17}.

\subsubsection{Connector}The Connector component bridges the front-end interface and back-end systems, ensuring seamless communication and real-time interaction. In this implementation, Flask acts as the back-end intermediary, transmitting user queries to the LLM via POST requests and retrieving generated responses. Flask formats the responses, incorporating both text and images, before delivering them back to the interface for user display. This architecture ensures efficient query processing and enhances system responsiveness.
\subsubsection{Database and Storage} This component manages structured storage for pre-processed data, including text and images. Organized in JSON format, this structure ensures quick and consistent retrieval for system queries.
\subsubsection{Process Manager} The Process Manager component oversees key tasks such as data pre-processing, keyword extraction, semantic matching, and response generation. It ensures the alignment of retrieved data with user requests.
\subsubsection{Display}The Display component provides the user interface for query input and response visualization, ensuring an intuitive and functional user experience.

\subsection{Data Collection and Pre-Processing}

The data collection process is implemented via an automated web scraping pipeline using Selenium \cite{b18}. This pipeline specifically targets 41 bridge-related technologies under the Bridge section on the InfoTechnology website \cite{b1}, systematically extracting textual descriptions and corresponding images. The scraped data is organized into a structured JSON format, enabling efficient access and streamlined processing for InfoTech Assistant operations. Sample data for the technologies "Hammer Sounding" and "Magnetic Particle Testing (MT)" \cite{b1} are illustrated in TABLE \ref{table: hammer} and TABLE \ref{table: mag}, respectively. This structured format allows for rapid retrieval and contextual presentation of information, supporting the system's ability to provide accurate and relevant responses based on user queries.

To enhance the dataset, additional infrastructure-related domains, including pavements, tunnels, and utilities, were identified for inclusion \cite{b0}. Publicly available datasets were scraped using Selenium \cite{b18} and preprocessed using a consistent pipeline aligned with the bridge data. The datasets were structured in JSON format, integrating textual and visual data to enable efficient querying. This expansion broadens the dataset's scope, equipping the InfoTech Assistant to handle diverse infrastructure-related queries effectively.

\subsection{Transformer Model}

The all-mpnet-base-v2 model \cite{b30}, developed by Microsoft, is a transformer-based architecture designed for semantic similarity tasks and integral to the InfoTech Assistant's language understanding capabilities \cite{b2}. Trained on over 1 billion sentence pairs, this compact 420 MB model combines efficient memory usage with robust performance across diverse domains. Its key features include mean pooling to summarize semantic content and normalized embeddings to improve reliability and accuracy in similarity scoring. These features make it well-suited for enabling the InfoTech Assistant to provide precise and contextually relevant responses \cite{b28}.

\subsection{Flask Integration}

Flask is a lightweight and flexible web framework \cite{b8} that facilitates real-time communication between the front end and back end in the InfoTech Assistant system. It routes user queries via HTTP requests to the model’s processing components, ensuring efficient response handling. Flask's scalability allows the system to maintain robust performance even under increasing loads. By integrating Flask with Retrieval-Augmented Generation (RAG) \cite{b21}, the application provides accurate, context-rich, and timely responses, making it a reliable choice for conversational AI implementations.

\subsection{LLM Model Integration}
\subsubsection{The Llama 3.1 Model}Llama 3.1, an advanced decoder-only transformer model \cite{b25}, forms the core of the InfoTech Assistant, enabling it to handle complex, context-aware queries. Its architecture incorporates cutting-edge training techniques such as supervised fine-tuning (SFT) \cite{b32} and direct preference optimization (DPO), which enhance data quality and response precision. The quantization from BF16 to FP8 ensures computational efficiency, enabling deployment on single-server nodes \cite{b33}. The model's capabilities, including a 128K context window and multi-turn conversation support, make it ideal for tasks requiring continuity and detailed, dynamic responses.

\subsubsection{The Mistral Model} The Mistral Model \cite{b23}, a lightweight and efficient transformer-based language model, also contributes to the InfoTech Assistant, enabling it to handle complex, context-aware queries. Its design emphasizes low-latency performance while maintaining high-quality responses. The compact architecture of Mistral allows for efficient deployment and processing, which is particularly advantageous for real-time applications.

\subsubsection{Temperature Parameters in Language Models}

A language model is characterized by numerous hyperparameters that govern its architecture, training process, and performance. Temperature is one such hyperparameter that modulates randomness in model outputs by scaling the logits before applying the softmax function \cite{b26}. The Temperature parameter \(T\) affects the probability distribution as follows\cite{b31}:

\begin{equation}
    P(x_i) = \frac{\exp(z_i / T)}{\sum \exp(z_j / T)}
\label{eq: temp}
\end{equation}

where \(z_i\) represents the logits, \(T\) is the temperature, and \(P(x_i)\) is the probability of token \(i\)  from Equation \ref{eq: temp}.

A low temperature (e.g., 0.1) produces deterministic outputs by emphasizing high-probability tokens, resulting in consistent but less diverse responses. A high temperature (e.g., 1.5) flattens the probability distribution, promoting greater variability and creativity in responses but at the cost of predictability.

The InfoTech Assistant employs a temperature setting of 0.7 to balance deterministic accuracy with conversational diversity \cite{b27}. This setting enables contextually relevant and adaptable responses, ensuring precise information retrieval for technical queries while maintaining a natural and engaging conversational tone.

\subsection{Evaluation Metrics}
\subsubsection{Cosine Similarity}
 In this "InfoTech Assistant" system, cosine similarity is used to evaluate the relevance of retrieved content to user queries, ensuring that responses closely match the semantic intent of the input\cite{b4}.
 
Cosine similarity is a mathematical measure used to determine the similarity between two non-zero vectors in an \(n\)-dimensional space \cite{b28}. It is widely used in natural language processing to compare the semantic similarity of vectorized text representations. The formula for cosine similarity is as follows:

\begin{equation}
    \text{Cosine Similarity} = \frac{\vec{A} \cdot \vec{B}}{\|\vec{A}\| \|\vec{B}\|}
\label{eq: cos}
\end{equation}

From Equation \ref{eq: cos} where \(\vec{A}\) and \(\vec{B}\) are the vectors being compared, \(\vec{A} \cdot \vec{B}\) is their dot product, and \(\|\vec{A}\|\) and \(\|\vec{B}\|\) are their magnitudes.

 Cosine similarity values range from \(-1\) to \(1\). A value of \(1\) indicates identical vectors, \(0\) indicates orthogonality (no similarity), and \(-1\) indicates complete dissimilarity.

Cosine similarity is scale-invariant, meaning it focuses on the orientation of the vectors rather than their magnitude, which makes it particularly suitable for comparing normalized text embeddings.

\subsubsection{Response Accuracy} Accuracy, in this study, is calculated based on a threshold applied to cosine similarity scores\cite{b4}. To determine accuracy, a predefined threshold (e.g., 0.85) is applied to cosine similarity scores. If the similarity score for a response meets or exceeds this threshold, the response is considered “correct”. 
Accuracy is computed as the ratio of correct responses to the total number of test cases, represented as a percentage as shown in Equation \ref{eq: Accuracy}.

\begin{equation}
\text{Accuracy (\%)} = \left( \frac{\text{Number of Correct Responses}}{\text{Total Number of Test Cases}} \right) \times 100
\label{eq: Accuracy}
\end{equation}

\section{Experimental Results and Analysis}

The pilot InfoTech Assistant integrates key components, including data collection, a user-friendly HTML interface \cite{b6}, and a state-of-the-art LLM. Automated web scraping extracts textual and visual data on 41 bridge technologies from the InfoTechnology web portal \cite{b1}, organizing it into a structured JSON format for efficient querying. The intuitive interface enables users to submit queries and receive detailed, image-enhanced responses, with Flask facilitating seamless communication between the front end and the LLM.

The performance of the InfoTech Assistant \cite{b34} was evaluated based on response accuracy, latency, and user satisfaction. The system underwent testing with both technical and non-technical users, and its performance metrics were analyzed over multiple testing rounds.

\begin{figure}[h]
\centerline{\includegraphics[width=0.75\linewidth]{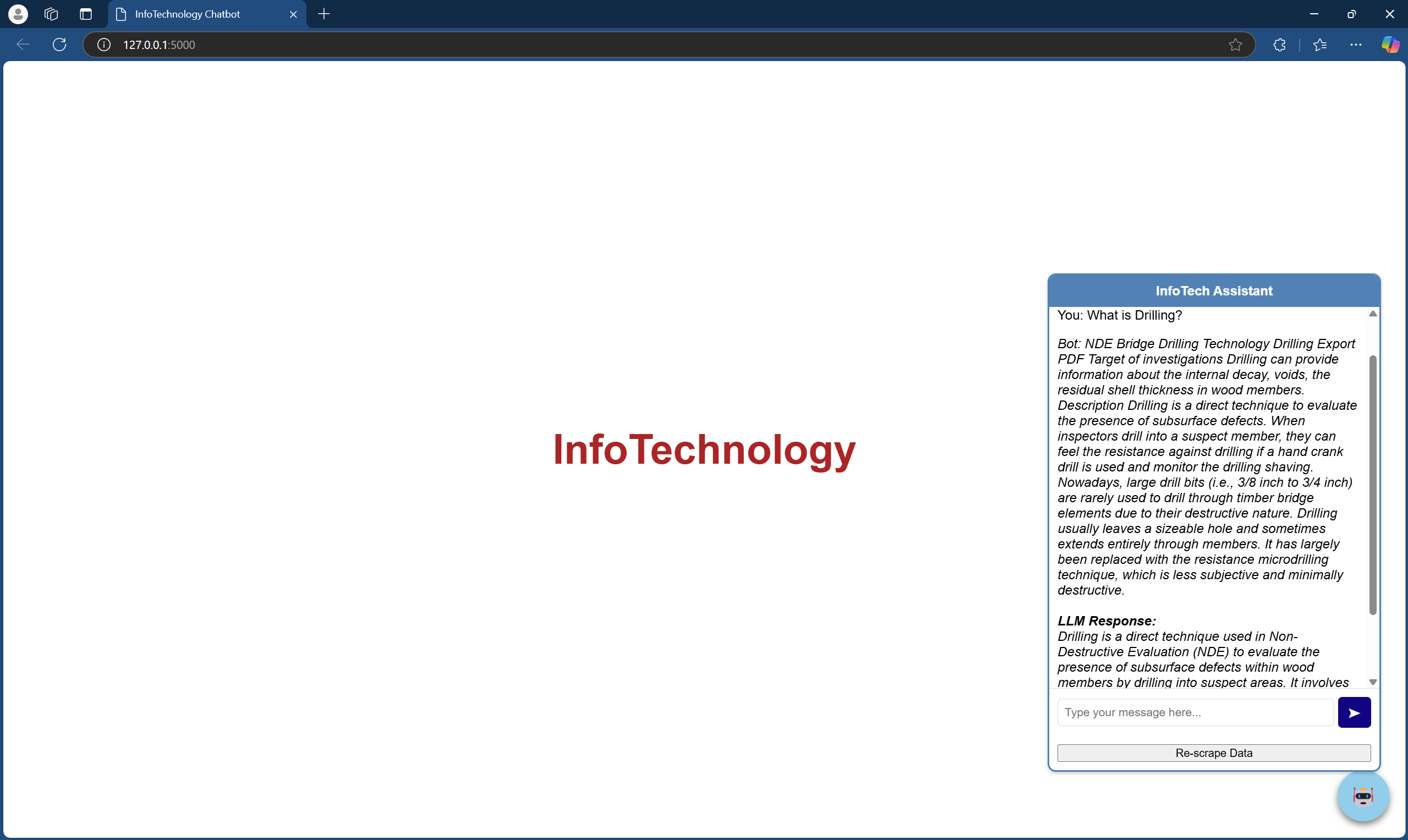}}
\caption{InfoTechnology Web Portal UI with InfoTech Assistant Interaction}
\label{fig: fig2}
\end{figure}

\subsection{Experiment Setup and System Requirements}
For the experiment, the Assistant system was deployed using an LM Studio server \cite{b3} hosting the Llama 3.1 8B model\cite{b25}, which was accessed through a locally run API. Once the LM Studio server is initiated, a front-end HTML page containing the InfoTech Assistant interface is launched. Users interact with the InfoTech Assistant by inputting queries through this interface. The server, running the large language model, processes these queries \cite{b24}, applies NLP techniques, and generates responses that are displayed in the chat interface as shown in Fig. \ref{fig: fig2}. Additionally, if the context contains relevant images, they are displayed alongside the text.

\subsection{Sample Responses and Image Retrieval Results}
Fig. \ref{fig: Fig3} illustrates sample interactions with the InfoTech Assistant, showcasing its capability to produce accurate, contextually relevant responses and retrieve related images. The InfoTech Assistant leverages RAG\cite{b7} to ensure fact-based answers are derived from the structured dataset, providing validated responses rather than generating content independently. Additionally, the system retrieves images relevant to user queries, thereby enriching the responses with visual context. For complex inquiries, the LLM further enhances user understanding by generating concise summaries\cite{b22}. Multi-image retrieval supports comprehensive insight, especially for technical topics requiring a visual aid for clarity.

\subsection{Comparison of Bot and LLM Responses}
In the InfoTech Assistant system, responses from the "Bot" and the "LLM" serve distinct but complementary roles. The Bot response is directly generated from data scraped from the official InfoTechnology website, ensuring the integrity and precision of the information provided. This approach allows the Bot response to present a comprehensive and detailed answer that closely reflects the technical content from the original source, thereby maintaining factual accuracy essential for professional use as shown in Fig. \ref{fig: Fig3} (a).

In contrast, the LLM response is a summarized version of the same information, crafted to enhance user comprehension and readability. By distilling the primary points, the LLM response provides an accessible summary that allows users to quickly capture essential insights\cite{b22}. This dual-response mechanism effectively addresses diverse user needs by combining detailed, source-based responses with a more concise, user-friendly summary as shown in Fig. \ref{fig: Fig3} (d).

\subsection{Evaluation and Analysis}

The performance of the InfoTech Assistant was evaluated using key metrics, focusing on response accuracy and contextual relevance. These metrics were derived from the Assistant’s ability to accurately retrieve content and deliver appropriate responses, including visual data when available.

\subsubsection{Contextual Relevance}: The InfoTech Assistant effectively managed contextually relevant queries, but exhibited limitations when addressing highly specific or nuanced queries \cite{b24}. For instance, ambiguous questions occasionally returned broadly related information instead of precise answers \cite{b19}. This highlights potential areas for improvement, such as enhanced fine-tuning or the adoption of a larger model \cite{b28}, to better handle complex contextual inquiries.

\begin{figure}[htbp]
\centering
    \begin{subfigure}[t]{0.48\linewidth}
        \centering
        \includegraphics[width=0.9\linewidth]{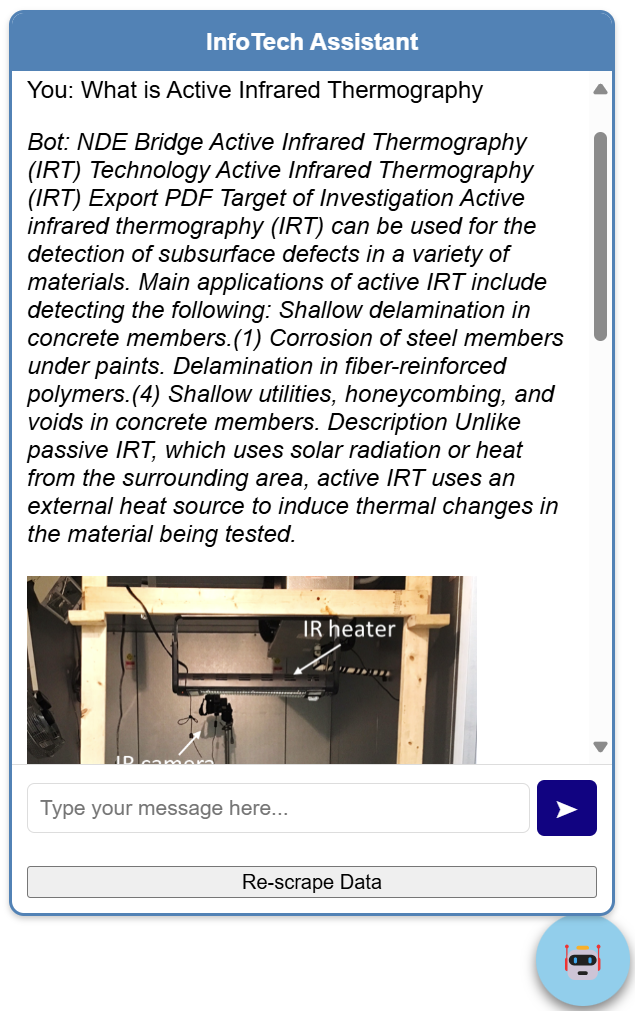} 
        \caption{User Query with Textual Response.}
        \label{fig: Fig3a}
    \end{subfigure}
    \begin{subfigure}[t]{0.48\linewidth}
        \centering
        \includegraphics[width=0.9\linewidth]{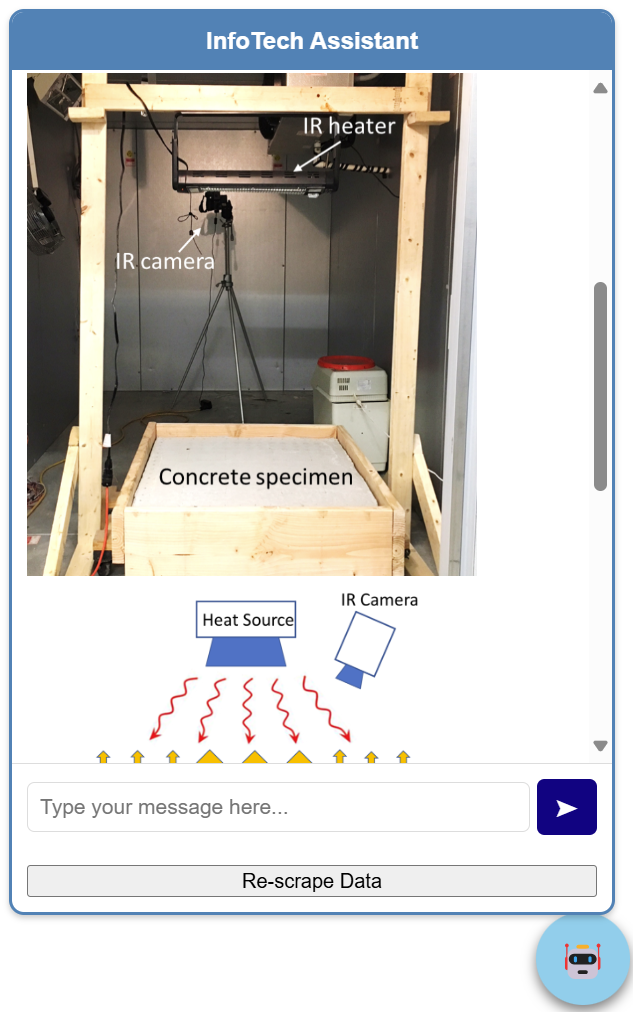} 
        \caption{User Query and Retrieved Image.}
        \label{fig}
    \end{subfigure}
    \begin{subfigure}[t]{0.48\linewidth}
        \centering
        \includegraphics[width=0.9\linewidth]{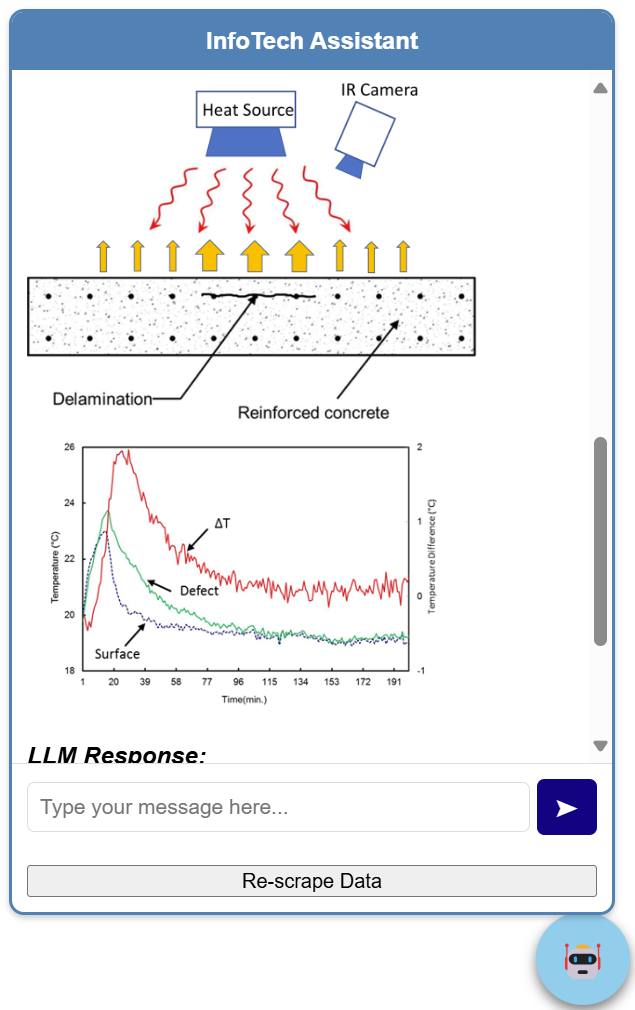} 
        \caption{Response with Multiple Images.}
        \label{fig}
    \end{subfigure}
    \begin{subfigure}[t]{0.48\linewidth}
        \centering
        \includegraphics[width=0.9\linewidth]{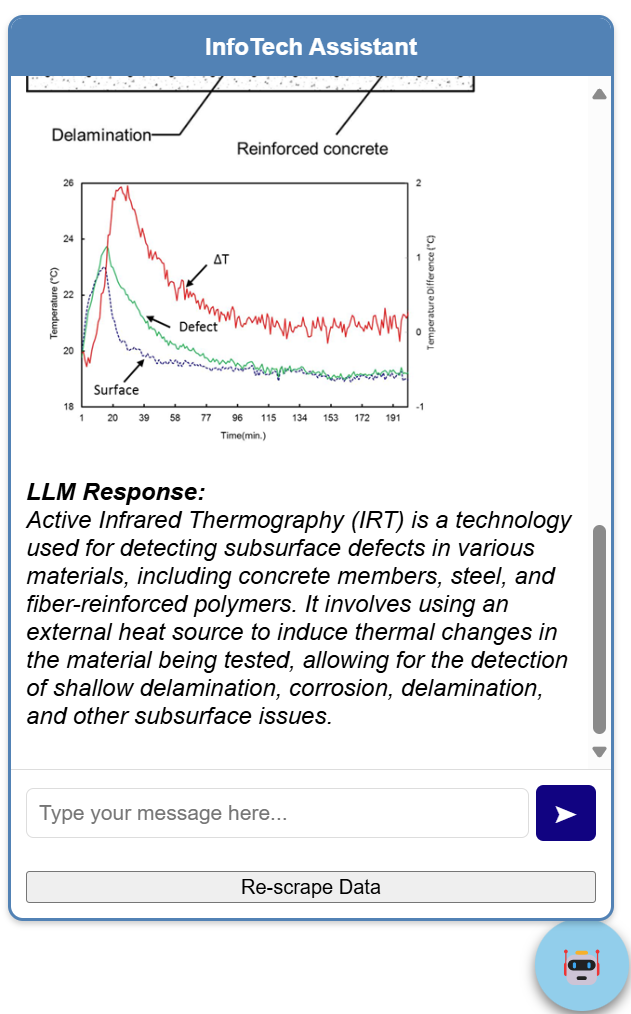} 
        \caption{Generated LLM Response.}
        \label{fig: Fig3d}
    \end{subfigure}
\caption{User Queries and Responses by the InfoTech Assistant.}
{Image Source: Adapted from\cite{b1}}
\label{fig: Fig3}
\end{figure}

\subsubsection{Latency and Scalability}

The latency of the InfoTech Assistant, defined as the response time from receiving a user query to delivering an answer, was measured during testing. The Llama 3.1 model exhibited a latency range of 15 to 20 seconds, primarily due to local processing demands. In comparison, the Mistral-7B-Instruct-v0.2 model demonstrated a latency range of 10 to 22 seconds, reflecting its efficiency in query processing. Latency was observed to be dependent on the system's computational capabilities, with increased processing power reducing response times \cite{b34}. The test system configuration is detailed in TABLE \ref{table: sysspec}.

\subsubsection{Similarity Calculation and Accuracy Evaluation}

The accuracy of the InfoTech Assistant was evaluated using cosine similarity, a widely used metric in natural language processing to measure semantic alignment between vector embeddings of text \cite{b29}. Cosine similarity values, calculated using Equation \ref{eq: Accuracy}, range from 0 to 1, with scores of 0.85 or higher considered correct.

Expected and actual responses were vectorized using the Sentence-Transformer model \cite{b2}, and accuracy was determined as the percentage of correct responses among the total test cases. As shown in TABLE \ref{table: Llama samples}, the Llama 3.1 model achieved similarity scores of 0.94 and 0.92, and accuracies of 95\% and 94\% for queries like "What is Electrical Resistivity" and "What are benefits of Hammer Sounding," respectively. Similarly, as presented in TABLE \ref{table: Mistral samples}, for the same queries  the Mistral 7B model achieved similarity scores of 0.90 and 0.92, with accuracies of 92\% and 94\%. These results demonstrate the Assistant's capability to provide semantically aligned and contextually accurate responses.

\begin{table}[h]
\centering
\caption{Sample Questions Similarity and Accuracy Results for Llama 3.1}
 \begin{tabular}
{|p{3.8cm}|p{1.1cm}|p{0.8cm}|p{1cm}|}
\hline
\textbf{Question} & \textbf{Similarity} & \textbf{Status} & \textbf{Accuracy} \\
\hline
What is Electrical Resistivity? & 0.94 & Correct & 95\% \\
\hline
What are benefits of Hammer Sounding? & 0.92 & Correct & 94\% \\
\hline
\end{tabular}
\label{table: Llama samples}
\end{table}

\begin{table}[h]
\centering
\caption{Sample Questions Similarity and Accuracy Results for Mistral 7B}
\begin{tabular}
{|p{3.8cm}|p{1.1cm}|p{0.8cm}|p{1cm}|}
\hline
\textbf{Question} & \textbf{Similarity} & \textbf{Status} & \textbf{Accuracy} \\
\hline
What is Electrical Resistivity? & 0.90 & Correct & 92\% \\
\hline
What are benefits of Hammer Sounding? & 0.92 & Correct & 94\% \\
\hline
\end{tabular}
\label{table: Mistral samples}
\end{table}

The comparison between the pre-trained models Llama 3.1 and Mistral 7B was conducted over 15 rounds of testing, using 15 different questions related to various technologies available on the InfoTechnology Bridge website. The similarity scores and overall accuracies for both models were calculated and analyzed.

\begin{figure}[h]
\centerline{\includegraphics[width=0.95\linewidth]{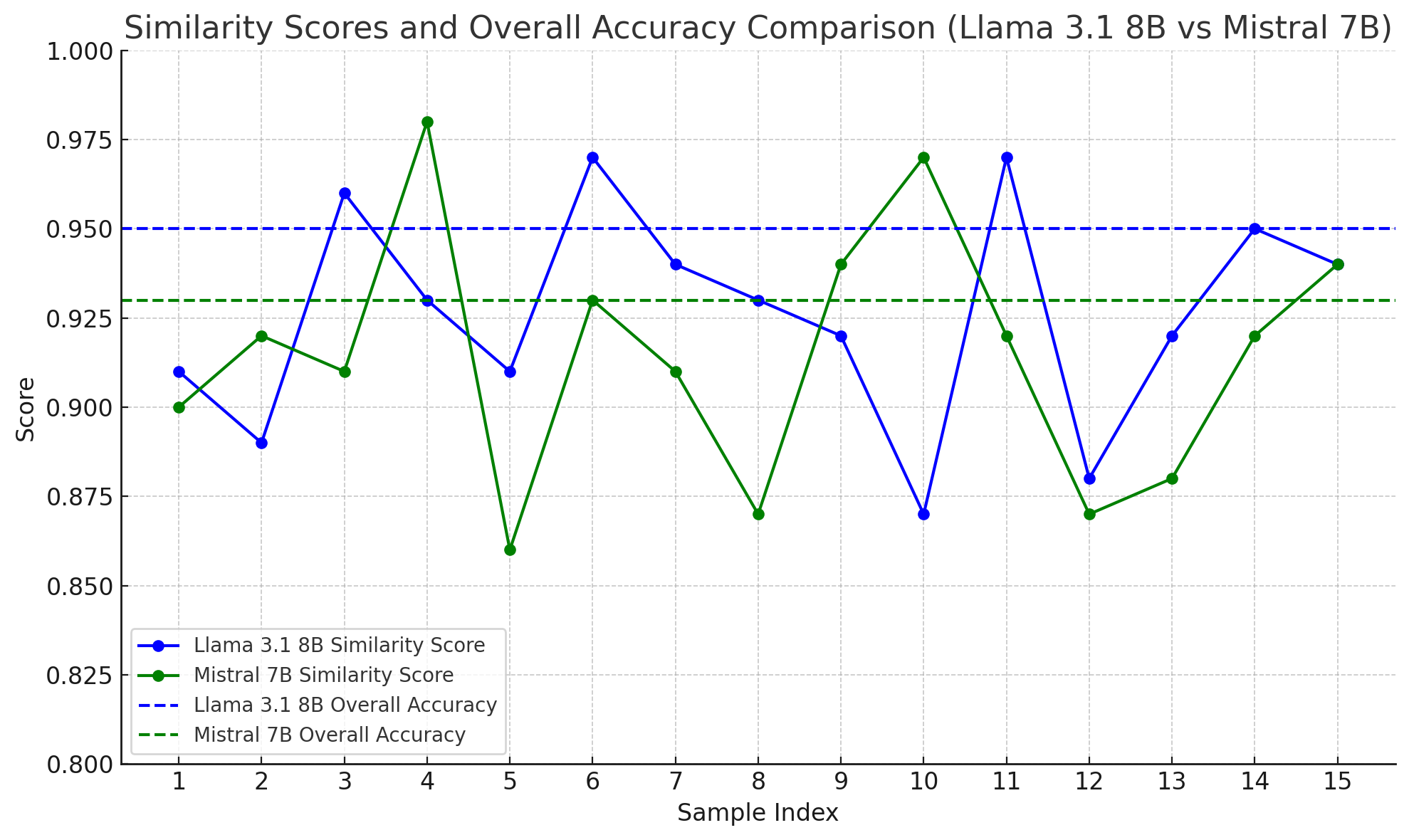}}
\caption{Similarity scores and overall accuracies of Llama 3.1 8B and Mistral 7B}
\label{fig: Fig4}
\end{figure}

Fig. \ref{fig: Fig4} illustrates the individual similarity scores for each sample, with overall accuracies represented by horizontal dashed lines. The Llama 3.1 model \cite{b25} achieves an overall accuracy of 95\%, which is marginally superior to the Mistral 7B model \cite{b23}, with an accuracy of 93\%. These visualizations effectively highlight the consistency and performance differences between the models, with Llama 3.1 8B demonstrating enhanced semantic alignment in comparison to Mistral 7B.

TABLE \ref{table: allsim} provides a comprehensive breakdown of similarity and accuracy scores for each question across both models. This detailed representation offers valuable insights into the granular performance of Llama 3.1 8B and Mistral 7B on specific tasks, complementing the observations presented in Fig. \ref{fig: Fig4}.

\section{Conclusions}

The InfoTech Assistant demonstrates the potential of conversational agents in addressing domain-specific challenges in infrastructure technology. By employing advanced techniques such as data scraping, Retrieval-Augmented Generation (RAG), and a large language model (LLM) hosted on LLM Studio, the system delivers accurate and contextually relevant information. Its architecture, comprising structured data collection, a Flask-based backend, and a user-friendly interface, enables efficient and precise responses to complex queries related to bridge technologies.

The integration of RAG significantly enhances the accuracy of LLM responses by grounding them in factual and domain-specific data, establishing the InfoTech Assistant as a reliable resource for infrastructure professionals. Evaluations validate its capability to manage diverse queries, offering both textual and visual outputs. Additionally, the use of pre-trained LLMs ensures versatility, allowing the system to provide detailed answers from a locally stored JSON database while leveraging a broader knowledge base for general queries.

In terms of performance, the Llama 3.1 8B model achieved an accuracy of 95\%, outperforming the Mistral 7B model, which recorded an accuracy of 93\%. However, the Mistral 7B model exhibited lower latency, highlighting a trade-off between accuracy and response time. Interestingly, the similarity metric showed limitations in fully evaluating the models' capabilities. For example, in certain cases, the Llama model provided correct and detailed answers but achieved a lower similarity score due to its tendency to include additional explanatory content. This finding suggests that while similarity can be indicative, it may not be the most reliable metric for assessing model performance in this context.

\section{Future Work}

Future enhancements for the InfoTech Assistant will prioritize reducing latency and improving its capability to manage multi-turn conversations effectively. Key advancements include the integration of advanced models such as Falcon 180B \cite{b36}, which are designed to deliver low-latency and contextually rich responses, alongside improvements to Retrieval-Augmented Generation (RAG) for more accurate and precise content retrieval. Additionally, the system will leverage domain-specific large language models (LLMs) and fine-tuning techniques \cite{b28} to enhance its performance and relevance in addressing technical queries. To support these advancements, the system will adopt cloud-based infrastructure to enable faster data access and scalable operations. A dynamic re-scraping mechanism will also be implemented to ensure that responses remain aligned with the most up-to-date information. These enhancements aim to significantly improve the system's contextual awareness, responsiveness, and overall user experience, solidifying its role as a reliable tool for infrastructure technology professionals.

\section*{Acknowledgment}

The authors would like to thank Meta and Mistral, the team behind the LLaMA model, for providing a powerful language tool that was instrumental in the development of the InfoTech Assistant. The authors also extend their gratitude to the contributors of the Transformers library, whose tools facilitated the seamless integration of the system. Additionally, the authors acknowledge the support of the Federal Highway Administration (FHWA) for granting access to the publicly available InfoTechnology web portal, the data from which was crucial for the training and testing of the model. These contributions were vital in making this project successful and effective.

\onecolumn

\appendix

\subsection*{Sample Data from InfoTech Assistant JSON file}

\begin{table}[h]
\centering
\caption{Supplementary Data Table: Scraped Data for Post ID 2769 - Hammer Sounding Technology}
\begin{tabular}{|p{3cm}|p{12cm}|}
\hline
\textbf{Field} & \textbf{Data} \\
\hline
\textbf{ID} & 2769 \\
\hline
\textbf{Summary} & NDE Bridge - Hammer Sounding Technology: Hammer sounding is beneficial for identifying shallow defects in wood structures and is sensitive to severe-stage defects. It involves tapping the wood surface with a hammer and listening for hollow or dull sounds to detect damaged areas. This technique is best suited for small regions, typically following visual inspection to confirm areas of suspected damage. \\
\hline
\textbf{Description} & Hammer sounding uses a hammer as an excitation source, while the inspector’s ears serve as receivers. When intact wood is struck, the sound frequency reflects its thickness and the sound velocity. For wood with near-surface defects, only the top layer above the defect vibrates, making it possible to distinguish between defective and intact areas by sound. \\
\hline
\textbf{Data Acquisition} & A pick hammer, often used by geologists, is recommended for this method. The flat end is used for sounding, while the pick end is suitable for probing. Hammer sounding is typically used on side-grain, with additional probing possible on side- and end-grain. For further analysis of detected defects, advanced methods like stress wave timing or resistance micro drilling can be applied. \\
\hline
\textbf{Data Processing} & No data processing is required. \\
\hline
\textbf{Data Interpretation} & Defective regions are marked by hollow or dull sounds. \\
\hline
\textbf{Advantages} & Hammer sounding is quick, simple, low-cost, and widely accepted in field inspections. \\
\hline
\textbf{Limitations} & The technique may not detect early or intermediate decay stages, is subjective, and lacks quantitative data on wood properties. \\
\hline
\textbf{References} & White, R. H., and R. J. Ross, eds. (2014). \textit{Wood and Timber Condition Assessment Manual.} 2nd ed., USDA Forest Service. \newline Ryan, T. W., et al. (2012). \textit{FHWA Bridge Inspector's Reference Manual (BIRM).} FHWA, Washington, DC. \\
\hline
\textbf{Images} & 
\begin{itemize}
    \item \url{https://infotechnology.fhwa.dot.gov/wp-content/uploads/2022/07/hammer-sounding.png}
\end{itemize} \\
\hline
\textbf{Text URL} & 
\begin{itemize}
    \item \url{https://infotechnology.fhwa.dot.gov/hammer-sounding/}
\end{itemize} \\
\hline
\end{tabular}
\label{table: hammer}
\end{table}

\begin{table}[h]
\centering
\caption{Supplementary Data Table: Scraped Data for Post ID 129 - Magnetic Particle Testing (MT) Technology}
\begin{tabular}{|p{3cm}|p{12cm}|}
\hline
\textbf{Field} & \textbf{Data} \\
\hline
\textbf{ID} & 129 \\
\hline
\textbf{Summary} & NDE Bridge - Magnetic Particle Testing (MT) Technology: MT is commonly used to detect cracks in steel girders, steel truss members, and other steel structures like sign supports and light poles. It can be applied during fabrication to ensure weld quality or to in-service structures to detect service-induced cracks. \\
\hline
\textbf{Description} & MT locates surface and subsurface discontinuities in ferromagnetic materials by applying a magnetic field to the material. If discontinuities exist, a leakage field forms, attracting ferromagnetic particles to outline the defect. Magnetic particles can be applied as dry particles or in a liquid carrier. \\
\hline
\textbf{Physical Principle} & The method works on magnetic induction and magnetic field leakage principles, applicable only to ferromagnetic materials. Defects cause local magnetic flux leakage, and fine magnetic particles align with the flux lines to highlight the disruption caused by defects. \\
\hline
\textbf{Data Acquisition} & MT uses a “dry powder” method, where magnetic particles with colored dye are applied to the material surface. Excess powder is removed, leaving particles held by the magnetic field to indicate discontinuities. Surface preparation involves removing coatings to avoid non-relevant indications. Orientation of magnetic fields is crucial, often requiring reorientation of equipment to ensure detection of all cracks. \\
\hline
\textbf{Data Processing} & MT requires no data processing as crack indications are detected by visual inspection. \\
\hline
\textbf{Data Interpretation} & MT indications are visually inspected to identify relevant signs of cracks. Indications from non-crack sources, such as surface contamination, are distinguished by the inspector, and relevant findings are documented. \\
\hline
\textbf{Advantages} & MT is low-cost, widely available, requires minimal data processing, and has simple result interpretation. \\
\hline
\textbf{Limitations} & It only detects surface or near-surface flaws, and surface preparation is needed. \\
\hline
\textbf{References} & ASTM, Standard Guide for Magnetic Particle Testing, E709-08, ASTM International, Conshohocken, PA, 2008. \\
\hline
\textbf{Images} & 
\begin{itemize}
    \item \url{https://infotechnology.fhwa.dot.gov/wp-content/uploads/2021/04/mt_1.jpg}
\end{itemize} \\
\hline
\textbf{Text URL} & 
\begin{itemize}
    \item \url{https://infotechnology.fhwa.dot.gov/magnetic-particle-testing-mt/}
\end{itemize} \\
\hline
\end{tabular}
\label{table: mag}
\end{table}

\vspace{6 cm}
\begin{figure*}[htbp]
\centering

\begin{minipage}{\textwidth}
\centering
\captionof{table}{System Specifications for Experimental Setup}
\label{table:sysreq}
\begin{tabular}{|p{6cm}|p{10cm}|}
\hline
\textbf{Specification} & \textbf{Details} \\
\hline
\textbf{Operating System} & Microsoft Windows 11 Home \\
\hline
\textbf{System Manufacturer} & Dell Inc. \\
\hline
\textbf{System Model} & Inspiron 16 Plus 7630 \\
\hline
\textbf{Processor} & 13th Gen Intel Core i7-13620H, 2.40 GHz \\
\hline
\textbf{Total Cores} & 10 Cores, 16 Logical Processors \\
\hline
\textbf{Installed RAM} & 32.0 GB (31.7 GB usable) \\
\hline
\textbf{Total Physical Memory} & 31.7 GB \\
\hline
\textbf{Available Physical Memory} & 8.77 GB \\
\hline
\textbf{Total Virtual Memory} & 44.3 GB \\
\hline
\textbf{Available Virtual Memory} & 8.94 GB \\
\hline
\textbf{System Type} & 64-bit Operating System, x64-based Processor \\
\hline
\end{tabular}
\label{table: sysspec}
\end{minipage}

\end{figure*}

\begin{table*}[h]
\centering
\caption{Similarity and Accuracy Scores of Sample Questions for Llama and Mistral Models}

\begin{tabular}{|c|p{9.25cm}|c|c|c|c|}
\hline
\textbf{No.} & \textbf{Questions} & \multicolumn{2}{c|}{\textbf{Llama}} & \multicolumn{2}{c|}{\textbf{Mistral}} \\ \hline
 &  & \textbf{Similarity} & \textbf{Accuracy} & \textbf{Similarity} & \textbf{Accuracy} \\ \hline
1 & What is NDE Drilling? & 0.91 & 0.93 & 0.90 & 0.92 \\ \hline
2 & How is Acoustic Tomography used in NDE bridge technology? & 0.89 & 0.91 & 0.92 & 0.94 \\ \hline
3 & What is meant by Automated Sounding? & 0.96 & 0.98 & 0.91 & 0.93 \\ \hline
4 & How to work with NDE Ground Penetrating Radar? & 0.93 & 0.95 & 0.98 & 0.99 \\ \hline
5 & What is Impact Echo (IE)? & 0.91 & 0.93 & 0.86 & 0.88 \\ \hline
6 & What is Ultrasonic Surface Waves (USW)? & 0.97 & 0.99 & 0.93 & 0.95 \\ \hline
7 & Can you explain how to do Screw Withdrawal Testing? & 0.94 & 0.96 & 0.91 & 0.93 \\ \hline
8 & What is NDE Radiography (RAD Tendons)? & 0.93 & 0.95 & 0.87 & 0.89 \\ \hline
9 & Explain Moisture Content Measurement? & 0.92 & 0.94 & 0.94 & 0.96 \\ \hline
10 & What is Magnetometer (MM)? & 0.87 & 0.89 & 0.97 & 0.99 \\ \hline
11 & How to do Magnetic Particle Testing (MT)? & 0.97 & 0.99 & 0.92 & 0.94 \\ \hline
12 & What is Infrared Thermography (IT)? & 0.88 & 0.90 & 0.87 & 0.89 \\ \hline
13 & What is meant by Half-Cell Potential (HCP)? & 0.92 & 0.94 & 0.88 & 0.90 \\ \hline
14 & Explain Galvanostatic Pulse Measurement (GPM)? & 0.95 & 0.97 & 0.92 & 0.94 \\ \hline
15 & What is Eddy Current Testing (ECT)? & 0.94 & 0.96 & 0.94 & 0.96 \\ \hline
\end{tabular}
\label{table: allsim}
\end{table*}

\end{document}